\documentclass[
]{ceurart}

\usepackage{listings}
\usepackage{algpseudocode}
\usepackage{algorithm}
\usepackage{amsmath}
\usepackage[shortlabels]{enumitem}
\usepackage[position=top]{subfig}

\newcommand{\class}{c}
\newcommand{\notclass}{\neg\class}

\usepackage{xcolor}

\begin{document}

\copyrightyear{2025}
\copyrightclause{Copyright for this paper by its authors.
  Use permitted under Creative Commons License Attribution 4.0
  International (CC BY 4.0).}

\conference{TRUST-AI: The European Workshop on Trustworthy AI. Organized as part of the European Conference of Artificial Intelligence - ECAI 2025. October 2025, Bologna, Italy.}

\title{Beyond single-model XAI: aggregating multi-model explanations for enhanced trustworthiness}

\author[1]{Ilaria Vascotto}[%
orcid=0009-0004-3961-1674,
email=ilaria.vascotto@phd.units.it,
]
\cormark[1]

\address[1]{Department of Mathematics, Informatics and Geosciences, University of Trieste, Trieste, Italy}

\author[1,2]{Alex Rodriguez}[%
orcid=0000-0002-0213-6695,
email=alejandro.rodriguezgarcia@units.it,
]

\address[2]{The Abdus Salam International Center for Theoretical Physics, Trieste, Italy}

\author[3]{Alessandro Bonaita}
[email = alessandro.bonaita@generali.com]

\address[3]{Assicurazioni Generali Spa, Milan, Italy}

\author[1]{Luca Bortolussi}[
orcid = 0000-0001-8874-4001,
email = lbortolussi@units.it, 
]

\cortext[1]{Corresponding author.}

\begin{abstract}
The use of Artificial Intelligence (AI) models in real-world and high-risk applications has intensified the discussion about their trustworthiness and ethical usage, from both a technical and a legislative perspective. The field of eXplainable Artificial Intelligence (XAI) addresses this challenge by proposing explanations that bring to light the decision-making processes of complex black-box models. Despite being an essential property, the robustness of explanations is often an overlooked aspect during development: only robust explanation methods can increase the trust in the system as a whole. This paper investigates the role of robustness through the usage of a feature importance aggregation derived from multiple models ($k$-nearest neighbours, random forest and neural networks). Preliminary results showcase the potential in increasing the trustworthiness of the application, while leveraging multiple model's predictive power.   

\end{abstract}



\maketitle

\section{Introduction}

The role of Machine Learning (ML) and Artificial Intelligence (AI) has become ever so prominent in the latest years. The widespread use of these tools, especially in high-risk applications \cite{pneumonia,finance_longbing}, has contributed to a growing discussion in the field over their ethical and fair usage. Despite being highly accurate and capable of dealing with complex problems, these models lack \textit{transparency}, an essential property to truly \textit{trust} the used systems. Recent legislative regulations, such as the GDPR \cite{gdpr} and the AI Act \cite{aiact} in the European Union, have stressed the need of model transparency and the relevance of explanations to aid both end-users and practitioners in understanding the mechanics of \textit{black box} models. In light of these considerations, the field of eXplainable Artificial Intelligence (XAI) proposes a wide range of explanation methods that aim at \textit{opening the black box} and clarifying the decision-making process of the models. While multiple approaches have been proposed in the latest years, with the aim of explaining the predictions either locally (for a single datapoint) or globally (for the whole model), the fundamental role of explanation robustness is often left unconsidered. Robustness can be broadly defined as the ability of an explanation method (or \textit{explainer}) to propose similar explanations for similar inputs. Its role is of upmost importance to ensure that the explanations can be trusted and, consequently, to increase the trustworthiness of the system. Only by trusting the given explanation it is possible to build trust into the model itself. Crucially, popular techniques such as LIME \cite{lime} and SHAP \cite{shap} have been proved to be non robust in multiple works \cite{fooling_lime_shap, alvarezmelis} but are still being used extensively even in practical high-risk applications. An additional issue that may arise in the field is the disagreement problem \cite{krishna-lakkaraju}, that is, the cases in which multiple explanation methods are applied to the same instance and contrasting explanations are retrieved. In this case, the lack of a ground truth for explanations makes it virtually impossible to choose one method over the other, effectively damaging the positive applications of XAI. 

In light of these considerations, we aim at tackling two research questions on explanation trustworthiness. The first research direction posits that the disagreement problem may be mitigated by considering an aggregation of multiple explanations. The second research direction instead aims at increasing system's trustworthiness by computing a local robustness score to determine whether an explanation can be trusted.
Previous attempts at answering these questions were made in \cite{vascotto}, where a robustness analysis of feature importance methods was proposed alongside an ensemble approach. The focus was on tabular datasets, classification problems and neural network models (NNs). Trustworthiness was investigated with respect to a robustness estimator computed on a neighbourhood of the datapoint, constructed by leveraging the manifold hypothesis. An aggregation of approaches tailored to NNs \cite{ig,deeplift,lrp} was introduced and it was shown that multiple neural networks - with varying architectures but similar accuracy - were able to detect when a datapoint would lie in a robust (explanation-wise) area of the dataspace and, in that case, would also agree on the prediction.

Taking inspiration from these results, we propose a natural generalization of the approach to multiple classes of models, aiming at answering both research questions. In particular, we investigated if an ensemble approach could be used even in cases in which multiple decision making processes are being considered at once, and tested the trustworthiness of the derived explanations. We have focused on $k$-nearest neighbours ($k$-NNs), random forests (RFs) and NNs. The choice of these three models was governed by the differences in which the models make a prediction, having $k$-NN a distance-base method and RF an ensemble of rule-based base learners, significantly differing with respect to NNs in terms of complexity, accuracy and inherent interpretability. As in \cite{vascotto}, the class of explanations which was selected is that of feature attributions, as they offer significant and interpretable results even to less experience users. A contribution in this area is the introduction of two new feature attribution approaches for the $k$NN and the RF, as both models tend to prefer explanations of other forms.

The remainder of this paper is structured as follows: Section \ref{methodology} will present the proposed  methodology, Section \ref{results} the preliminary results on tabular datasets and binary classification tasks and Section \ref{conclusions} future research directions.

\section{Methodology}\label{methodology}

\subsection{Explaining k-Nearest Neighbours}\label{section:knn}
Despite being considered an \textit{interpretable} method, $k$-NNs still lack a proper feature importance approach to explain their decisional process. In particular, while it is easy to understand the distance-based reasoning when predicting a new datapoint's class, it is not as simple to identify which features were the most influential for the given prediction.

Leveraging its inner mechanisms, we have developed a new feature importance approach to locally explain the decisional process of $k$-NNs. Having selected a value for the hyperparameter $k$, a new datapoint is classified based on the most frequent class which is encountered among its $k$ nearest neighbours within the training set. The difference between the two possible classes can mainly be attributed to the features that exhibit larger distances between the two sets of corresponding points. This reasoning can be further extended when considering a point for which all the $k$-nn are associated to the same class. Locally, there are no distinctions to be emphasised, but the classification can be imputed to the fact that the datapoint is closer to those of the predicted class rather than the opposing one. This concept aligns with the reasoning of $k$-NN on a greater scale: datapoints are classified based on the most similar class, and the similarity is defined with distance-based metrics.

Our proposal makes use of this characteristics and produces a feature-importance explanation as described in Algorithm \ref{knn_explanation}. Consider two sets of datapoints, namely $N_\class$ and $N_{\notclass}$, which are comprised of the $k_e$ nearest neighbours selected, respectively, between the training set's datapoints belonging to class $\class$ and those of the other class $\notclass$. The average feature-wise distance within each set of points and the datapoint of interest $\mathbf{x}$ is computed and stored as $D_\class$ and $D_{\notclass}$. Assuming $\mathbf{x}$ is predicted to belong to class $\class$, the explanation is proposed as the feature-wise difference $e = D_{\notclass} - D_\class$. To ensure comparability, the resulting vector is normalized to have norm one.

\begin{algorithm}[t]
    \caption{$k$-NN explanation: $\mathbf{x}$ is the input point and $f$ is the $k$-nearest neighbours model}
    \begin{algorithmic}
    \Function{average\_distance}{$\mathbf{x}, N$}
    \State $D \gets 1/\vert N\vert \cdot \sum_{\mathbf{x}'\in N} dist(\mathbf{x}, \mathbf{x}')$
    
    \Return $D$
    \EndFunction
    \State

        \Function{Explain\_knn}{$\mathbf{x}, k_e$}      
        \State $\class=f(\mathbf{x})$ \Comment Class predicted by the $k$-nn model
        \State $N_\class \gets \Call{k-nn}{k_e, \mathbf{x} \vert \class}$ \Comment $k_e$ nearest neighbours belonging to class $\class$
        \State $N_{\notclass} \gets \Call{k-nn}{k_e, \mathbf{x} \vert \notclass}$

       \State $D_\class \gets \Call{average\_distance}{\mathbf{x}, N_\class}$ \Comment Average feature-wise distance
       \State $D_{\notclass} \gets \Call{average\_distance}{\mathbf{x}, N_{\notclass}}$

        \State $e \gets D_{\notclass} - D_\class$
        \State $e \gets e / \vert\vert e\vert\vert_2$ \Comment Normalize to have norm 1
        
        \Return $e$
    \EndFunction

    \end{algorithmic}
    \label{knn_explanation}
\end{algorithm}

\begin{algorithm}[b]
    \caption{RF explanation: $\mathbf{x}$ is the input point and $f$ is the random forest model}
    \begin{algorithmic}

        \Function{sum\_node\_impurity}{path, $e$}
        \For {$\text{node} \in \text{path}$}
            \State $i \gets \text{node.feature}$
            \State $e[i] \gets \Call{add}{\text{node.impurity}}$
        \EndFor
        
        \Return $e$
        \EndFunction
\State
        \Function{Explain\_rf}{$\mathbf{x}, f$}
        \State $p_\class, p_{\notclass} \gets f(\mathbf{x})$ \Comment Predicted class probabilities
        \State $\class \gets \arg\max(f(\mathbf{x}))$ \Comment Predicted class

        \State $ e_\class, e_{\notclass}  \gets \text{array}(0, \text{dim = }n_{ft})$
        
        \For {$\text{tree} \in f$} 
            \State $\text{path} \gets \Call{decision\_path}{\text{tree}, \mathbf{x}}$ \Comment Follow the traversed path in the tree
         \If {$\text{tree.predict}(\mathbf{x}) = \class$}
         \State $e_\class \gets \Call{sum\_node\_impurity}{\text{path}, e_\class}$
        \Else 
        \State $e_{\notclass} \gets \Call{sum\_node\_impurity}{\text{path}, e_{\notclass}}$
         \EndIf
        \EndFor
        \State $e \gets (p_{\notclass}+\epsilon)\times e_\class - p_\class \times e_{\notclass}$
                \State $e \gets e / \vert\vert e\vert\vert_2$ \Comment Normalize to have norm 1

        \Return $e$
        \EndFunction
    \end{algorithmic}
    \label{rf_explanation}
\end{algorithm}

\subsection{Explaining Random Forests}\label{section:rf}
Random Forest explanations are usually proposed in the form of single decision trees or sets of rules, to align with the binary decisions that occur at each node split. While explanations of this form can be useful, it is hard to compare them to feature attributions derived from other classes of models. For this reason, we have developed a new feature attribution method for random forests (presented in Algorithm \ref{rf_explanation}), based on the computed node impurity at each node of the traversed paths in the forest.

Consider a datapoint $\mathbf{x}$ which is passed through a random forest $f$ composed of $B$ trees. Assume that the random forest predicts point $\mathbf{x}$ belonging to class $\class$, having predicted probabilities $[p_\class, p_{\notclass}]$. For each of the $B$ trees in the forest, the datapoint's prediction is made by traversing the tree following a single decision path. Depending on whether the individual tree's prediction aligns with the random forest majority vote, the decision path includes features that contribute to the RF predicted class or to the opposite one. By construction, each node in a decision tree is selected as the one with minimal node impurity, practically measured by the Gini index or the cross entropy. We propose to use the node impurities as feature importances, with due adjustments, considering their relevance in the training phase of the model itself. Consider two feature attribution vectors, namely $e_\class$ and $e_{\notclass}$, of length equal to the number of features in the input vector $\mathbf{x}$. For each tree in the forest, we retrieve the decision path followed by point $\mathbf{x}$ and check if the prediction of the individual tree is the same as that of the random forest or not. For a tree which agrees with the RF prediction, we retrieve the feature importance vector $e_\class$ and, for all the features used at the split nodes of the decision path, sum the node impurity to the corresponding element of $e_\class$. The vector $e_{\notclass}$ is otherwise used to store the node impurities in a similar manner. A final calibration is performed to properly take into account the effect of positively and negatively influencing features. In particular, the feature importance coefficients defined above are weighted by the opposite class probability scores, returning a vector of the form $e = (p_{\notclass}+\epsilon)\cdot e_\class - p_\class \cdot e_{\notclass}$. The parameter $\epsilon=0.01$ yields non-null coefficients in the case in which $p_{\notclass}=0.00$. As in the $k$-NN case, a final normalization step is performed to ensure comparability.

\subsection{Aggregating Multi-Model Explanations} \label{ame}
As mentioned in the Introduction, we propose an aggregation of explanations derived from different models. While new approaches have been proposed in the previous subparagraphs, for NNs we will be using DeepLIFT \cite{deeplift} as explanation method. DeepLIFT is a local feature importance approach tailored to neural networks which is based on a backpropagation-like procedure to distribute the \textit{difference-from-reference} in the output layer to the input one. We have selected it as, during preliminary experiments, it exhibited a more consistent behaviour with respect to other approaches applicable to NNs. It is necessary to normalize its attributions to have norm $1$ for comparability with the other feature importance vectors before computing the aggregation.

Assume that the feature importance vectors for $k$-NN (Algorithm \ref{knn_explanation}), RF (Algorithm \ref{rf_explanation}) and NN (DeepLIFT \cite{deeplift}) are denoted, with no loss of generality, by $\mathbf{a}_l$ with $l\in\{\ 1, \dots, L\}, \, L=3$. In this study, we propose the aggregation to be computed as the feature-wise arithmetic average of the composing feature attributions vectors, namely:
\begin{equation}
    \mathbf{a}_{agg}=\cfrac{1}{L}\sum_{l=1}^{L} \mathbf{a}_l
\end{equation}

Among the advantages of using the average as an aggregation is its ability to penalize strong sign discordances between the individual methods, while still producing attributions with both positive and negative signs. It also takes into account instances in which one of the methods is \textit{uncertain} about the importance of a feature $j$, that is, when it is associated to a coefficient $\vert a_l^{(j)} \vert < \delta, \, (\delta>0)$.

\subsection{Robustness Estimation and Neighbourhood Generation} \label{rob_neigh}
Robustness estimation is a critical aspect of this research and, while multiple works have introduced robustness metrics for explanations, we selected the estimator proposed in \cite{vascotto}. The robustness estimate $\hat{\mathcal{R}}$ for a datapoint $\mathbf{x}$, a neighbourhood $\mathcal{N}$, an explanation method $e$ and a model $f$ is computed as:
\begin{equation}\label{robustness}
    \hat{\mathcal{R}}(\mathbf{x}, \mathcal{N}, e, f) = \cfrac{1}{\vert\mathcal{N}\vert}\sum_{\tilde{\mathbf{x}}\in\mathcal{N}} \rho(e(\mathbf{x}), e(\tilde{\mathbf{x}}))  
\end{equation}
where $\mathcal{N} = \{\tilde{\mathbf{x}} \vert \tilde{\mathbf{x}} = \mathbf{x} + \lambda \text{ with } \lambda \in \mathbb{R}^m, dist(\mathbf{x}, \tilde{\mathbf{x}}) < \epsilon \,(\epsilon>0) \, \text{ and } f(\mathbf{x})=f(\tilde{\mathbf{x}}) \}$, $m$ is the number of features and $\rho$ is the Spearman's rho rank correlation coefficient.

In literature, there is a clear distinction between perturbation-based robustness and adversarial robustness. The first one refers to the ability of an explainer to produce similar explanations for similar inputs under the expected data distribution. The latter one instead refers to malicious attacks to either the model or the explanation, and encompasses the ability of an explainer to produce similar outputs under such attacks. Our research focuses on non-adversarial perturbations and, as shown in \cite{vascotto}, we note that the neighbourhood generation mechanism can be highly influential in the computation of the robustness, even when considering perturbation-based evaluations. We suggest the use of the medoid-based neighbourhood which can be summarized by the following steps. First perform $k$-medoid clustering on a validation set: for each cluster, the medoid $\mathbf{x}^c$ acts as a representative. For each medoid, the set $NN^c$ stores the $k_M$ nearest neighbours computed among the other cluster centres. For each datapoint we'd like to perturb, the corresponding cluster is computed and both $\mathbf{x}^c$ and $NN^c$ are retrieved. A medoid $\mathbf{x}^M$ is selected at random from the set $NN^c$. Having $\alpha$ and $\alpha_{cat}$ the probabilities of perturbing numerical ($x_{num}$) and categorical ($x_{cat}$) variables respectively, a perturbation is then performed as:

\begin{equation}\label{neighbourhood}
    \begin{cases}
    \tilde{x}_{num} = (1-\bar \alpha)\cdot x_{num} + \bar\alpha \cdot x_{num}^M \text{ with } \bar\alpha \leftarrow Beta(\alpha\cdot100, (1-\alpha)\cdot 100)\\
    \tilde{x}_{cat} = \begin{cases}
        x_{cat} \text{ with probability } 1-\alpha_{cat} \\
        x_{cat}^M \text{ with probability } \alpha_{cat}
    \end{cases}
    \end{cases}
\end{equation}

This scheme yields perturbations which are on-manifold and therefore more truthful to the observed data distribution on which the model was trained. A final step is performed to remove the perturbations that change predicted class label with respect to the original point, ensuring comparability explanation-wise.

As we are considering three different models being used at the same time, the requirement that the predictions should be the same for both the original datapoint and its perturbation has to be satisfied for all three models concurrently. More specifically, $f_l(\mathbf{x}) = f_l(\tilde{\mathbf{x}}) \, \forall l\in(1, \dots, L)$.

\section{Preliminary Results}\label{results}

\paragraph{Dataset and Model Details}
We have tested our proposal on five publicly-available tabular datasets addressing binary classification problems\footnote{\url{https://archive.ics.uci.edu/datasets}}. Some preprocessing steps were performed on the data, such as the standardization of numerical variables, the one-hot encoding of categorical ones as well as the removal of highly correlated features. The dataset were split into a training, a validation and a test sets. For all the use-cases, we have trained a $k$-NN classifier, a random forest and a neural network. The $k$-NN models were trained with a number of neighbours set at $k=15$ for the adult and bank dataset and $k=5$ for the other cases. The random forest were built with $B=25$ base learners for all cases. All models exhibit an adequate performance as the accuracy scores are above the 80\% mark, as summarized in Table \ref{tab:dataset}, with the exception of the $k$-NN model on the heloc dataset. As can be expected, the accuracy of the $k$-NN models is often the lowest while the neural networks showcase larger values in the most complex datasets (adult, bank, heloc) and the random forest in the two simpler ones (cancer, wine).

\begin{table}[h]
\caption{Dataset details (number of numerical and categorical variables - after preprocessing - and train, validation, test sizes) and model accuracy over the training set. Values in bold indicate the highest accuracy per dataset.}
\label{tab:dataset}
\begin{tabular}{c|ccccc|ccc|}
\cline{2-9}
 & \multicolumn{5}{c|}{Dataset Details} & \multicolumn{3}{c|}{Model Accuracy - Train (\%)} \\ \hline
\multicolumn{1}{|c|}{Dataset} & \multicolumn{1}{c|}{\# Num} & \multicolumn{1}{c|}{\# Cat} & \multicolumn{1}{c|}{Train} & \multicolumn{1}{c|}{Valid} & Test & \multicolumn{1}{c|}{$k$-NN} & \multicolumn{1}{c|}{RF} & NN \\ \hline
\multicolumn{1}{|c|}{adult} & \multicolumn{1}{c|}{5} & \multicolumn{1}{c|}{7} & \multicolumn{1}{c|}{36177} & \multicolumn{1}{c|}{8045} & 1000 & \multicolumn{1}{c|}{81.45} & \multicolumn{1}{c|}{84.85} & \textbf{91.39} \\ \hline
\multicolumn{1}{|c|}{bank} & \multicolumn{1}{c|}{5} & \multicolumn{1}{c|}{9} & \multicolumn{1}{c|}{36168} & \multicolumn{1}{c|}{8043} & 1000 & \multicolumn{1}{c|}{88.57} & \multicolumn{1}{c|}{91.58} & \textbf{91.99} \\ \hline
\multicolumn{1}{|c|}{heloc} & \multicolumn{1}{c|}{14} & \multicolumn{1}{c|}{2} & \multicolumn{1}{c|}{8367} & \multicolumn{1}{c|}{1592} & 500 & \multicolumn{1}{c|}{75.51} & \multicolumn{1}{c|}{80.36} & \textbf{85.50} \\ \hline
\multicolumn{1}{|c|}{cancer} & \multicolumn{1}{c|}{15} & \multicolumn{1}{c|}{0} & \multicolumn{1}{c|}{397} & \multicolumn{1}{c|}{121} & 50 & \multicolumn{1}{c|}{95.71} & \multicolumn{1}{c|}{\textbf{99.75}} & 93.41 \\ \hline
\multicolumn{1}{|c|}{white wine} & \multicolumn{1}{c|}{9} & \multicolumn{1}{c|}{0} & \multicolumn{1}{c|}{3918} & \multicolumn{1}{c|}{780} & 200 & \multicolumn{1}{c|}{86.63} & \multicolumn{1}{c|}{\textbf{94.69}} & 89.23 \\ \hline
\end{tabular}
\end{table}

\paragraph{Neighbourhood Generation and Aggregation Computation} For each point in the test set, a neighbourhood was generated with hyperparameters selected such that, for the three models, at least $95\%$ of the generated datapoints were kept on average. Default choices for the hyperparameters are $\theta = \{k_M=5, \alpha = 0.05, \alpha_{cat}=0.05\}$. For the individual models the filtering is based on the constraint $f(\mathbf{x})=f(\tilde{\mathbf{x}})$ while for the aggregation it holds that $f_l(\mathbf{x})=f_l(\tilde{\mathbf{x}})\,  \forall l \in (1,\dots, L)$. 

Note that for the aggregation, the condition includes both the case in which all models agree on the prediction and that in which one of the three models predicts the opposing class. In this case, to correctly identify the feature importance vectors expressing the decision behind the models' predictions, it is possible to leverage the binary nature of the problems and change sign to the explanations associated to the disagreeing model before computing the aggregation. It is essential that the three explanations are referring to the same output class, to ensure comparability.

Let $f_l$ be a model associated to a feature importance vector $\mathbf{a}_l$ with $l\in(1,\dots, L)$. Assume that, for a given datapoint $\mathbf{x}$, it holds that $f_1(\mathbf{x})\neq f_2(\mathbf{x}) , f_2(\mathbf{x})=f_3(\mathbf{x})$. Then, we will consider as feature attributions the vectors of the form $(-\mathbf{a}_1,\mathbf{a}_2, \mathbf{a}_3)$ when computing the aggregation method $\mathbf{a}_{agg}$. 

\begin{figure}[t]
    \centering
    \includegraphics[width=.8\linewidth]{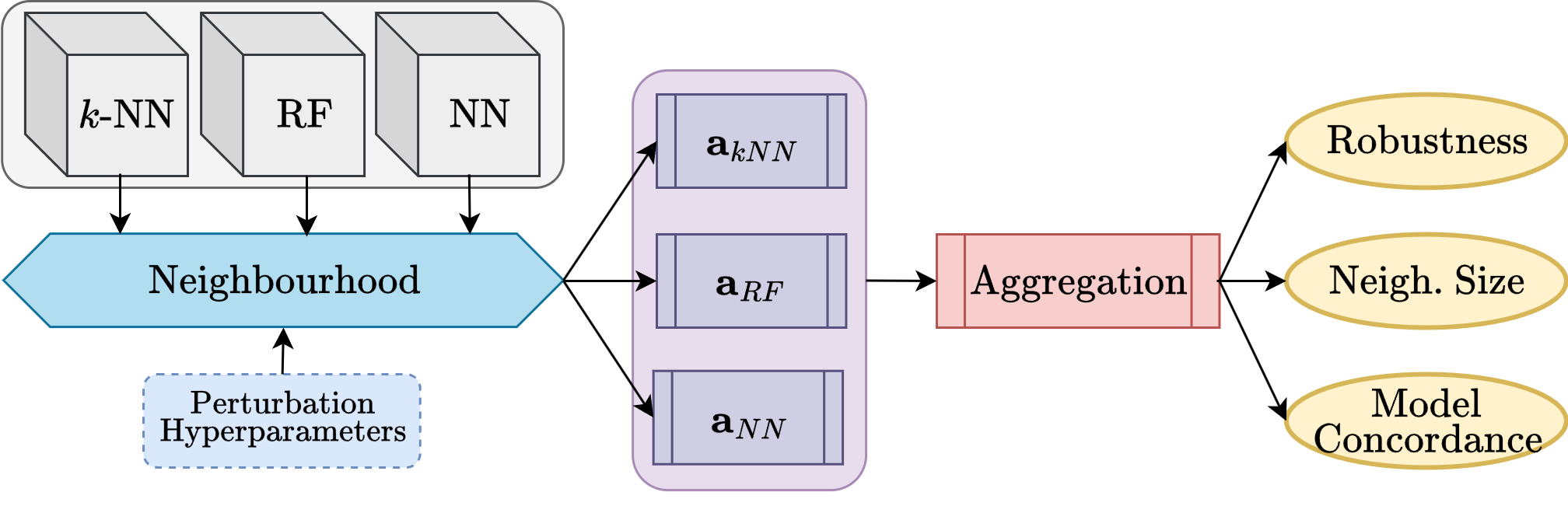}
    \caption{A summary of the methodology.}
    \label{fig:pipeline}
\end{figure}

For each model $f_l$ and corresponding explanation method $a_l$ then, the feature importances for the test set and the neighbourhoods are retrieved. The aggregation is performed following Section \ref{ame} taking into account the possible sign change for some attributions, as discussed. The robustness scores are then derived for the three individual model-explanation pairs and the aggregation, following Eq. \ref{robustness}. The complete framework is summarized in Figure \ref{fig:pipeline}.

\begin{figure}[b]
    \centering
    \includegraphics[width=\linewidth]{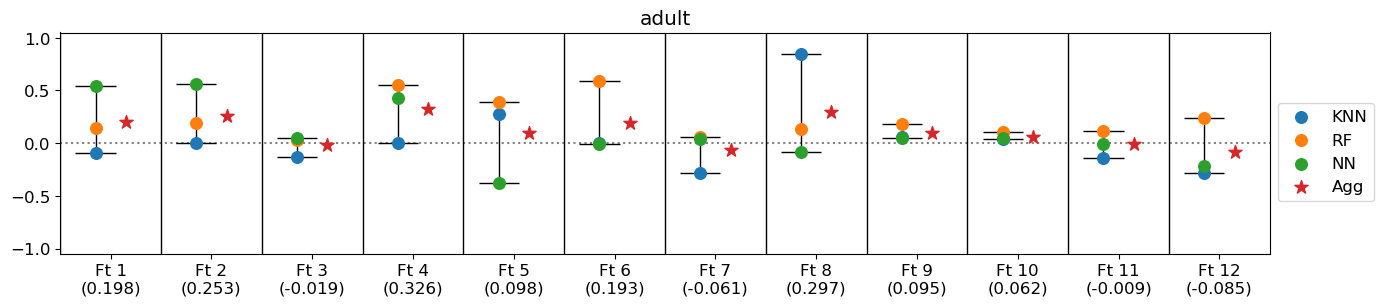}
    \caption{An example of the feature attributions on the adult dataset. The kNN (blue), RF (orange) and NN (green) explanations are connected by a vertical black line (the observed range). The red star represents the aggregation, whose value is presented within brackets close to the corresponding feature index (from 1 to 12).}
    \label{fig:agg}
\end{figure}

\paragraph{Qualitative Evaluation of the Aggregation} Let us briefly discuss an example of the derived aggregation, analysing the explanations of one datapoint $\mathbf{x}$ from the test set of adult dataset. The results are depicted in Figure \ref{fig:agg} where, for each of the twelve features, the three individual explanations (coloured circles) and the aggregation (red star) are presented. This example illustrates two key aspects of the aggregation: first, cases in which the three explanations have coefficients very close to zero (and possibly differing in sign) are shrunk towards the zero (that is, the feature is considered non important) - Features 3, 9, 10, 11. Secondly, when features are differing in sign and the magnitudes are large, the aggregation penalizes the disagreement but produces a coefficient which is coherent with the stronger signals - Features 1, 2, 4, 6, 8, 12. This is an expected (and desirable) behaviour as consistent information is kept and disagreements are penalized more strongly according to the magnitude.

\begin{figure}[h]
    \centering
    \includegraphics[width=0.90\linewidth]{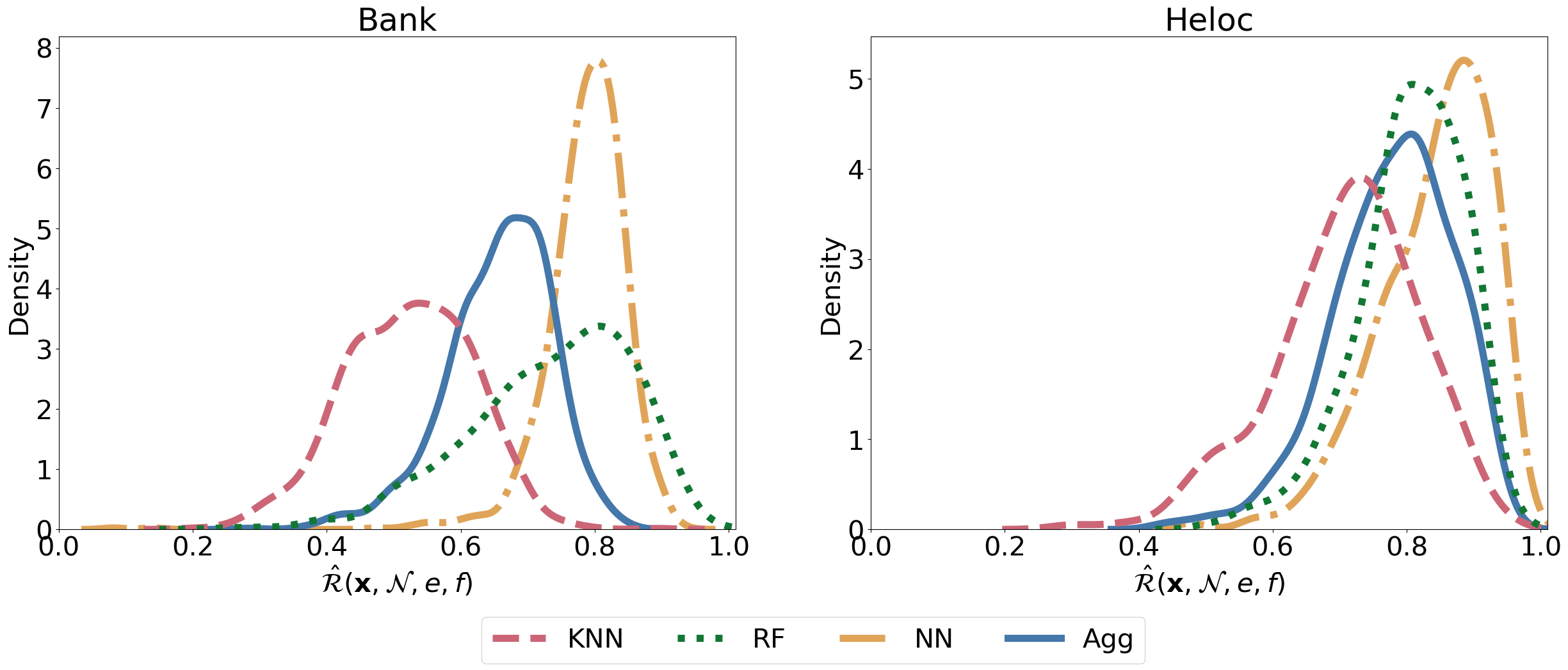}
    \caption{Distribution of the robustness scores for the three models and their aggregation on the bank (left) and heloc dataset (right).}
    \label{fig:robustness}
\end{figure}

\paragraph{Robustness Estimation}
Table \ref{tab:rob} presents the average robustness scores computed on the test set for the individual models and the aggregation. As can be seen also from Figure \ref{fig:robustness}, $k$-NN appears to be the least robust method on average, while NN the most. The aggregated explanation acts as an average - in terms of robustness - between the three individual components. This is to be expected as the robustness score was shown in \cite{vascotto} to be upper bounded by the individual methods' robustness. In fact, the advantage of using the aggregation is bifold. On one hand, the aggregation's robustness is still acceptable but, one the other, the aggregation encompasses the explanations derived from multiple explanations. In practice, it acts as a conservative and trustworthy explainer.

\begin{table}[h]
\caption{Average robustness scores (in percentage) over the whole test set. Values in bold indicate the highest robustness achieved within each dataset. }
\centering
\begin{tabular}{|c|c|c|c|c|}
\hline
Dataset & $k$-NN & RF& NN & Agg.  \\ \hline
adult & 61.12 & \textbf{88.67} & 85.03 & 74.58 \\ \hline
bank & 52.27 & 73.52 & \textbf{78.74} & 65.75 \\ \hline
heloc & 71.01 & 80.56 & \textbf{84.23 }& 77.92 \\ \hline
cancer & 83.31& 81.07 & \textbf{98.40} & 84.93 \\ \hline
white wine & 69.55 & 66.60 & \textbf{92.96} & 66.74 \\ \hline
\end{tabular}
\label{tab:rob}
\end{table}

Additionally, it is possible to note that the robustness scores derived from $k$-NN and RF align with both the model's inner reasoning and the explanation methods we have described in Subsections \ref{section:knn} and \ref{section:rf}. For the $k$-NN explanations, we expect the robustness to be lower on average as, by construction, the explanation method deviates the most from a local neighbourhood of the datapoint of interest. In particular, as we are always considering the distance to the $k_e$ nearest neighbours of class $\notclass$, it may be that - especially when the datapoint is far from the observed class boundary - the variability between the distances is higher. This reflects into lower robustness scores even if it locally mimics the decisional reasoning of the underlying model. On the other hand, RF exhibits greater robustness scores as it is harder to generate small perturbations that sensibly change the decision path traversed in each tree in the forest, maintaining lower explanation variability on average. 

\paragraph{Model Concordance and Neighbourhood Size} An interesting information which is easily derivable in this setting is the neighbourhood size as an additional metric of robustness. As for the aggregation multiple predictions are being considered at once, the local agreement between the models can be used as further confirmation of the robustness. The assumption is that, as suggested by our previous results, a robust explanation usually lies in an area of the data manifold which is robust explanation-wise. This can be verified by assessing if multiple models flag the same datapoint as being robust or not and are agreeing in the point's prediction, effectively verifying that the explanations are referred to the same class. In this context, the local agreement prediction-wise can indicate if the manifold area is robust also from an explanation point of view. 

\begin{table}[h]
\caption{Comparison between the models prediction agreement (\# obs) and average neighbourhood size in percentage (after filtering step). The columns Test presents the results over the whole test set, while the column Agree over the datapoints for which the three models agree on the prediction. The remaining columns are named by the model which disagrees on the prediction. Values in bold in each columns for the percentage indicate the largest neighbourhood size achieved per dataset.}

\centering
\begin{tabular}{c|cc|cc|cc|cc|cc|}
\cline{2-11}
 & \multicolumn{2}{c|}{Test} & \multicolumn{2}{c|}{Agree} & \multicolumn{2}{c|}{$k$-NN} & \multicolumn{2}{c|}{RF} & \multicolumn{2}{c|}{NN} \\ \hline
\multicolumn{1}{|c|}{Dataset} & \multicolumn{1}{c|}{\# obs} & $\%$ & \multicolumn{1}{c|}{\# obs} & $\%$ & \multicolumn{1}{c|}{\# obs} & $\%$ & \multicolumn{1}{c|}{\# obs} & $\%$ & \multicolumn{1}{c|}{\# obs} & $\%$ \\ \hline
\multicolumn{1}{|c|}{adult} & \multicolumn{1}{c|}{1000} & 77.41 & \multicolumn{1}{c|}{802} & \textbf{77.78} & \multicolumn{1}{c|}{100} & 75.86 & \multicolumn{1}{c|}{35} & 73.80 & \multicolumn{1}{c|}{63} & 77.14 \\ \hline
\multicolumn{1}{|c|}{bank} & \multicolumn{1}{c|}{1000} & 89.16 & \multicolumn{1}{c|}{929} & \textbf{91.03} & \multicolumn{1}{c|}{30} & 70.30 & \multicolumn{1}{c|}{10} & 49.10 & \multicolumn{1}{c|}{31} & 64.35 \\ \hline
\multicolumn{1}{|c|}{heloc} & \multicolumn{1}{c|}{500} & 67.23 & \multicolumn{1}{c|}{358} & 67.72 & \multicolumn{1}{c|}{70} & 61.27 & \multicolumn{1}{c|}{24} & \textbf{73.25} & \multicolumn{1}{c|}{48} & 69.25 \\ \hline
\multicolumn{1}{|c|}{cancer} & \multicolumn{1}{c|}{50} & 84.02 & \multicolumn{1}{c|}{46} & \textbf{89.61} & \multicolumn{1}{c|}{0} & N/A & \multicolumn{1}{c|}{3} & 25.67 & \multicolumn{1}{c|}{1} & 2.00 \\ \hline
\multicolumn{1}{|c|}{white wine} & \multicolumn{1}{c|}{200} & 78.04 & \multicolumn{1}{c|}{176} & \textbf{81.97} & \multicolumn{1}{c|}{12} & 56.50 & \multicolumn{1}{c|}{6} & 39.33 & \multicolumn{1}{c|}{6} & 44.67 \\ \hline
\end{tabular}
\label{tab:neigh}
\end{table}

Results from Table \ref{tab:neigh} show the average size retained in the neighbourhoods associated to the aggregation, starting from $n=100$ datapoints and after filtering out the perturbations that change class label with respect to the original one. It is possible to note that the average neighbourhood size varies greatly depending on the dataset being analysed (column Test - $\%$). The remaining columns of the table show the number of observations in test set and corresponding average neighbourhood size for the datapoints over which the three models agree on the prediction (Agree) and those for which one model (which names the column) predicts the opposite class with respect to the other two. Values in bold in the columns $\%$ are associated with the largest observed neighbourhood size after the filtering step. It can be noted that, in most cases, the largest size is achieved under the full concordance of the three models, effectively supporting our assumption. Note that, for the cancer dataset, there are $0$ datapoints for which $k$-NN is the disagreeing model, therefore the corresponding value N/A indicates that no data was available for the specific scenario during our experimental evaluation.

\paragraph{Validation Assessment} The results from Table \ref{tab:neigh} suggest that, even when considering multiple models of different types, it is possible to discuss the relationship between explanation robustness and model agreement, as we have previously observed in \cite{vascotto}. To verify our assumption, we have employed a validation assessment akin to the one presented in \cite{vascotto}. More specifically, we have considered a True Positive Rate (TPR) and a False Positive Rate (FPR) defined as:
    \begin{equation}
        \text{TPR} = \cfrac{\#\{\text{Robust \& Agree}\}}{\#\{\text{Agree}\}} \hspace{1cm} \text{FPR}=\cfrac{\#\{\text{Robust \& Disagree}\}}{\#\{\text{Disagree}\}}
    \end{equation}
where the agreement and disagreement are defined according to the model's concordance in the prediction and the robustness is defined, according to a threshold $r_{th}$, as $\mathbb{I}[\hat{\mathcal{R}}(\mathbf{x})\geq r_{th}]$.

\begin{figure}
    \centering
    \includegraphics[width=0.90\linewidth]{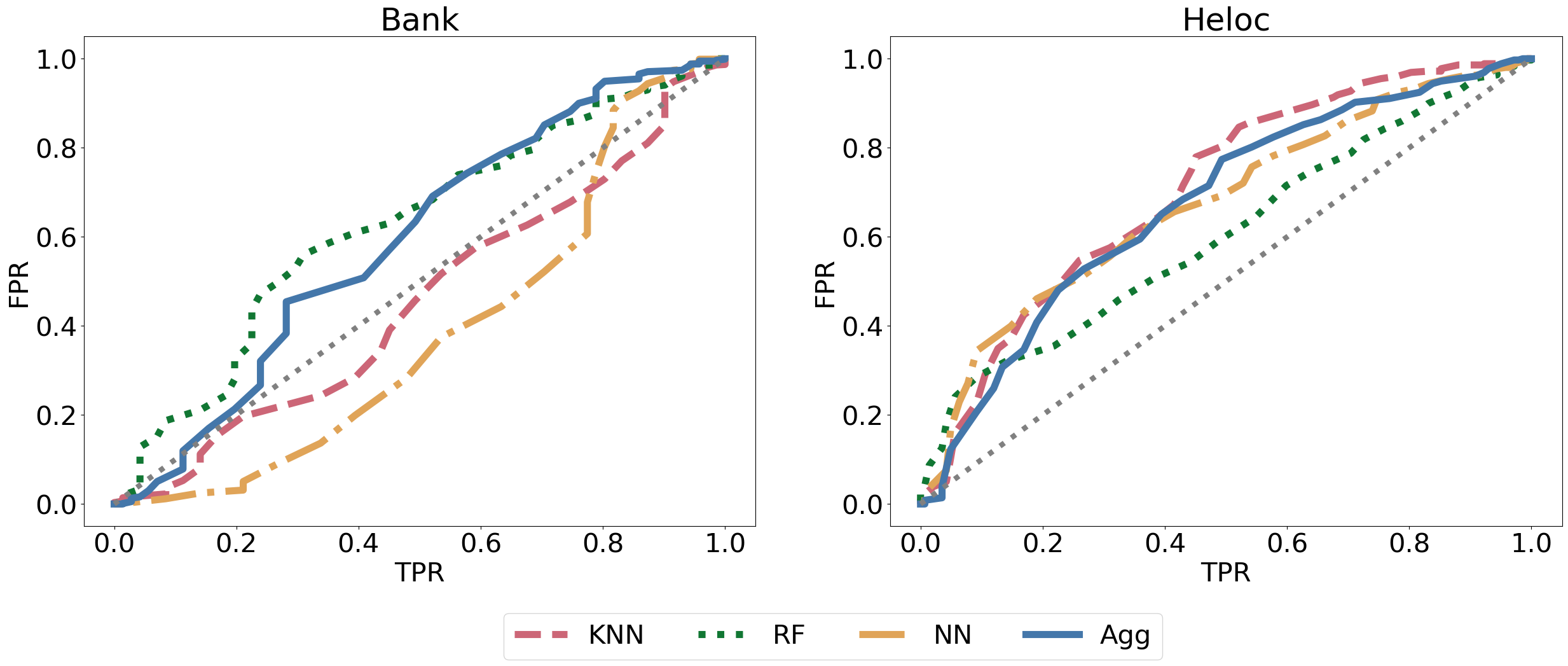}
    \caption{ROC curve computed for the three models and the aggregation for the bank (left) and heloc (right) datasets. The dotted line represents the bisecting line.}
    \label{fig:roc}
\end{figure}

Varying the threshold value, it is possible to obtain a ROC curve as in Figure \ref{fig:roc}, where the dotted gray line represents the bisecting line. The Area Under the Curve (AUC) can be used as a goodness measure: its values are reported in Table \ref{tab:auc}. The greater the value of the AUC, the more preferable is the given model-explanation pair under the agreement assumption and the considered dataset. The results from both Figure \ref{fig:roc} and Table \ref{tab:auc} suggest that the aggregation can act as a good explanation but not all models are performing in the expected manner. In particular, it seems that the $k$-NN attributions are the least effective in the selected scenarios.

\begin{table}[]
\caption{AUC values computed for the three models and the aggregation. Values in bold indicate the largest score recorded per dataset.}
\label{tab:auc}
\begin{tabular}{|c|c|c|c|c|}
\hline
Dataset & $k$-NN & RF & NN & Agg \\ \hline
adult & 0.4480 & 0.5417 & \textbf{0.6970} & 0.5901 \\ \hline
bank & 0.4128 & \textbf{0.6257} & 0.3861 & 0.6097 \\ \hline
heloc & 0.6573 & 0.6049 & \textbf{0.6748} & 0.6095 \\ \hline
cancer & 0.8397 & \textbf{0.9212} & 0.7120 & 0.9212 \\ \hline
white wine & \textbf{0.5088} & 0.4698 & 0.0469 & 0.4951 \\ \hline
\end{tabular}
\end{table}


\section{Conclusions and Future Work} \label{conclusions}
What we have presented in this contribution are the first steps towards a multi-model and multi-explanation aggregation to increase trustworthiness. The aim of our work is proposing an aggregation of explanations and models that can increase the trustworthiness in the system, by leveraging the predictive power of multiple models and their difference in the reasoning process. The robustness score and associated neighbourhood size indicate if the aggregated explanations lie in a robust area of the feature space and can therefore be trusted. Being a work in progress, our approach could still benefit from improvements on the following points:
\begin{itemize}
    \item An aggregation approach more complex than the arithmetic mean, able to deal with cases in which, for example, the attributions are of the form $(\nu, \nu, -2\nu)$ with $\nu>0$. The current aggregation would results in an attribution equal to $0$, compared to a magnitude (in absolute value) of at least $\nu$. Taking this aspect into account could benefit the reliability of the explanations.
    \item A new approach for $k$-NN explanations, as the proposed one is dependant on the hyperparameter $k_e$ and - even if from a theoretical point of view is consistent with the $k$-NN reasoning - it exhibits too large a variability.
    \item A more complete validation assessment of the proposal, to test its potential in real use-cases from both an explanation quality and trustability point of view. This would also allow to jointly consider the predictive power of the composing models more appropriately.
    \item The extension of the approach to other classes of models and different XAI approaches. This analysis would allow for an extensive evaluation of the efficacy of ensemble approaches under varying circumstances.
\end{itemize}


\begin{acknowledgments}
We wish to thank Assicurazioni Generali Spa for their support and interest in our work.
\end{acknowledgments}

\section*{Declaration on Generative AI}
  The authors have not employed any Generative AI tools.
  

\bibliography{bibliography}


\appendix
\renewcommand{\thefigure}{A\arabic{figure}}
\setcounter{figure}{0}

\renewcommand{\thetable}{A\arabic{figure}}
\setcounter{table}{0}

\section{Comparison with LIME and SHAP}
\begin{figure}[h]
    \centering
    \includegraphics[width=0.90\linewidth]{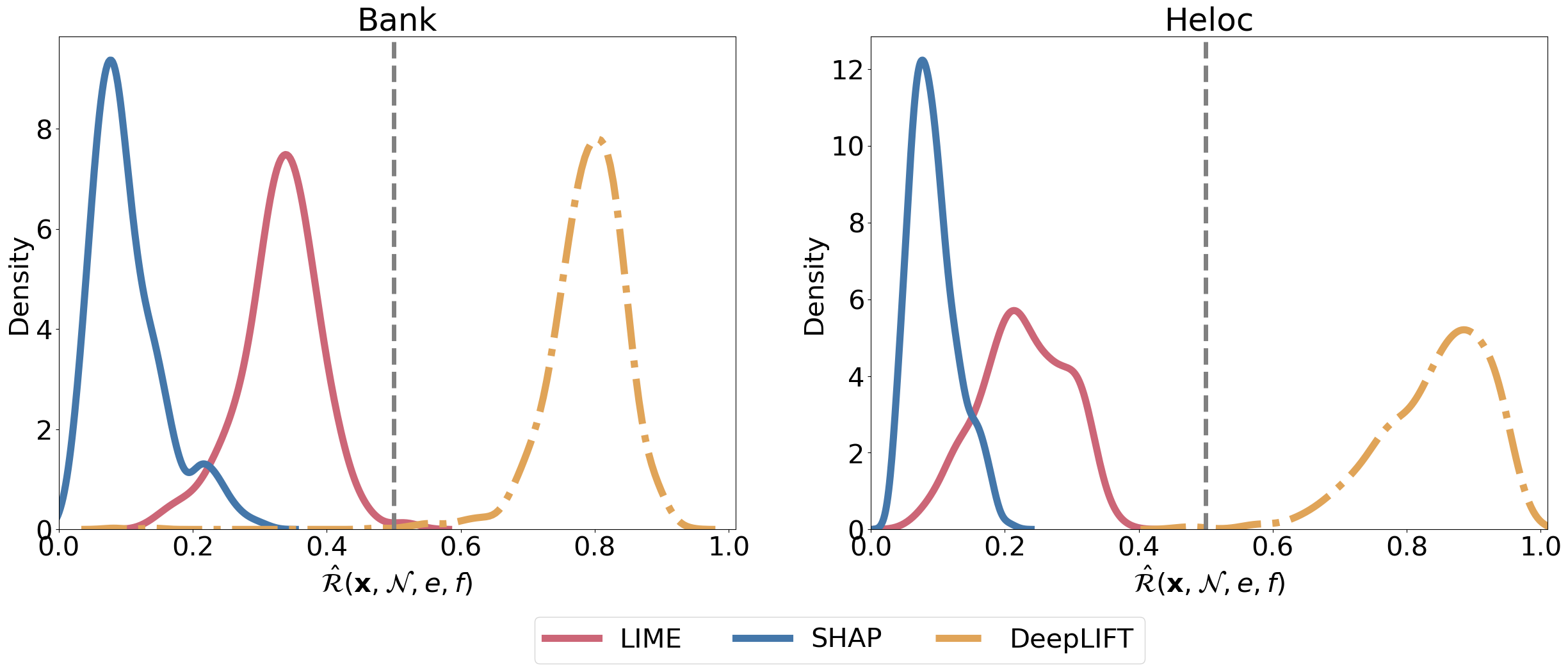}
    \caption{Distribution of the robustness scores for LIME, SHAP and DeepLIFT on the bank (left) and heloc dataset (right). All three XAI methods were applied to the outputs of the neural network model for comparability. The gray dashed line represents a robustness threshold of $0.50$.}
    \label{fig:limeshap}
\end{figure}

This appendix presents a comparative example of LIME and SHAP's robustness. Figure \ref{fig:limeshap} depicts the robustness scores of LIME, SHAP and DeepLIFT for the bank and heloc dataset (as in Figure \ref{fig:robustness}), with the yellow dotted-dashed line representing DeepLIFT's robustness in both figures. For comparability, LIME and SHAP (which are model-agnostic methods by nature) have been applied to the same model in this example - the neural network. It is clear from Figure \ref{fig:limeshap} that both LIME and SHAP are \textit{unstable} methods as their robustness scores are, on average, much lower than the DeepLIFT counterpart. Critically, they lie below the $0.50$ threshold (gray dashed line), denoting high instability.

Table \ref{tab:lime_shap} presents the average robustness scores derived on all the considered datasets for LIME, SHAP and DeepLIFT over the neural network. Note that the column "DeepLIFT" corresponds to the column "NN" of Table \ref{tab:rob}. Again, it is clear that the robustness scores of LIME and SHAP underperforms with respect to other techniques. 

In light of these considerations, as well as further support from the literature, LIME and SHAP were excluded from our analysis as their inner instability may damage both the individual explanations and any ensemble approach that may include them. 

\begin{table}[h]
\caption{Average robustness scores (in percentage) over the whole test set. All three XAI methods were applied to the outputs of the neural network model for comparability.}
\label{tab:lime_shap}
\begin{tabular}{|c|c|c|c|}
\hline
Dataset & LIME & SHAP & DeepLIFT \\ \hline
adult & 17.77 & 13.50 & 85.03 \\ \hline
bank & 33.11 & 10.29 & 78.74 \\ \hline
heloc & 22.80 & 9.34 & 84.23 \\ \hline
cancer & 56.11 & 17.33 & 98.40 \\ \hline
white wine & 53.99 & 36.08 & 92.96 \\ \hline
\end{tabular}
\end{table}

\end{document}